# A Decoupled Human-in-the-Loop System for Controlled Autonomy in Agentic Workflows

Edward C. Cheng
echeng04@stanford.edu

Jeshua Cheng
jeshua.cheng@inquiryon.com

***Abstract*** – *AI agents are increasingly deployed to execute tasks and make decisions within agentic workflows, introducing new requirements for safe and controlled autonomy. Prior work has established the importance of human oversight for ensuring transparency, accountability, and trustworthiness in such systems. However, existing implementations of Human-in-the-Loop (HITL) mechanisms are typically embedded within application logic, limiting reuse, consistency, and scalability across multi-agent environments.*

*This paper presents a decoupled HITL system architecture that treats human oversight as an independent system component within the agent operating environment. The proposed design separates human interaction management from application workflows through explicit interfaces and a structured execution model. In addition, a design framework is introduced to formalize HITL integration along four dimensions: intervention conditions, role resolution, interaction semantics, and communication channel. This framework enables selective and context-aware human involvement while maintaining system-level consistency.*

*The approach supports alignment with emerging agent communication protocols, allowing HITL to be implemented as a protocol-level concern. By externalizing HITL and structuring its integration, the system provides a foundation for scalable governance and progressive autonomy in agentic workflows.*

## I. BACKGROUND

The emergence of AI agents marks a transition from response-oriented systems to action-oriented systems capable of executing tasks, making decisions, and interacting with external environments. While this expanded capability enables powerful automation, it also introduces new risks, including incorrect execution, bias propagation, and misalignment with user intent. In high-stakes domains such as finance, healthcare, and enterprise operations, human involvement remains essential to ensure contextual judgment, accountability, and alignment with organizational policies.

Human-in-the-Loop (HITL) has therefore become a widely adopted paradigm for integrating human oversight into AI systems. In practice, HITL spans multiple stages of the AI lifecycle, including data curation, model training, decision validation, and post-deployment governance. However, despite its broad adoption, HITL is often realized through a collection of disjoint techniques, such as active learning, reinforcement learning from human feedback (RLHF), and application-specific approval workflows. As observed in prior studies, this fragmented treatment reflects the absence of a unified system-level framework for incorporating human oversight across heterogeneous AI systems [1].

This fragmentation becomes particularly problematic as AI systems evolve into autonomous agents. In many implementations, human approval logic is tightly coupled with application workflows, leading to duplicated logic, inconsistent behavior, and limited scalability across multiple agents. More importantly, existing approaches focus primarily on how to obtain human feedback, rather than how to systematically manage human involvement as a primary component of the system. This gap highlights a fundamental challenge: the need to operationalize HITL not merely as an interaction pattern, but as a structured, scalable, and governable subsystem.

In prior work, we introduced the Three-Pillar Model (3PM) for safe AI agent operation, emphasizing transparency, accountability, and trustworthiness as the foundation for progressive autonomy [2]. Building on this framework, this paper proposes treating HITL as an independent, decoupled system component within agent architectures. By defining explicit interfaces, policies, and execution semantics for human involvement, the proposed approach enables consistent, auditable, and scalable integration of HITL across agentic workflows, providing a practical foundation for controlled and progressive autonomy.

## II. FROM AI-IN-THE-LOOP TO HUMAN-IN-THE-LOOP

Recent work by Natarajan et al. has highlighted an important distinction in systems that combine human expertise and machine intelligence. Many systems commonly described as Human-in-the-Loop (HITL) are more accurately characterized as AI-in-the-Loop (AI2L), in which humans remain the primary decision-makers and AI components provide supporting information, recommendations, or analysis [3]. In these systems, the overall process is centered on human judgment, and the role of AI is to improve efficiency or decision quality without assuming control over outcomes.

This distinction is not merely semantic. It reflects a fundamental difference in system design, control, and evaluation. In AI2L systems, decision authority resides with the human, and the system is evaluated primarily based on its impact on human performance, including usability, interpretability, and decision quality. In contrast, HITL systems are typically structured such that the AI component drives the decision process while selectively incorporating human input for handling ambiguity, correction, supervision, or guidance. As a result, HITL systems are often evaluated using model-centric metrics such as accuracy or convergence, whereas AI2L systems require human-centered evaluation criteria that account for interaction quality and downstream outcomes.

The difference in control structure also affects how risks are introduced and managed. In AI2L systems, risks are associated with how humans interpret and act on AI-generated outputs, including issues such as over-reliance or misinterpretation. In HITL systems, risks are more directly tied to the behavior of the generative AI model itself, including incorrect actions, bias propagation, and failures under uncertainty. These differences imply that the design of oversight mechanisms must be aligned with the underlying control model.

As AI systems evolve toward agentic workflows, the boundary between AI2L and HITL becomes increasingly dynamic. Early-stage systems often operate in an AI2L configuration, where humans retain full control and AI provides assistance in perception, reasoning, or recommendation. However, as agents acquire the ability to execute actions autonomously, decision authority begins to shift toward the agent. In this setting, human involvement is no longer continuous but selective, and is typically triggered by risk, uncertainty, ambiguity, or policy constraints.

This transition introduces a hybrid operating mode in which AI agents perform the majority of routine decisions, while humans are engaged for exceptions, high-risk scenarios, or cases requiring contextual judgment. The design objective therefore shifts from maximizing automation to managing collaboration between human and agent. In particular, the system must determine when human intervention is required, how responsibility is assigned, and how human input is incorporated into the decision process.

A key implication of this shift is that human involvement can no longer be treated as an implicit or ad hoc feature of individual applications. Instead, it must be explicitly modeled and systematically managed as part of the agent operating environment. Without such structure, increasing autonomy leads to inconsistent behavior, unclear responsibility boundaries, and difficulty in enforcing governance policies across workflows.

This motivates the need for a more formal treatment of Human-in-the-Loop mechanisms in agentic systems. In the following section, we examine how HITL can be designed as an independent system component, enabling consistent and scalable integration of human oversight across applications.

| Dimension | AI-in-the-Loop (AI2L) | Human-in-the-Loop (HITL) |
| --- | --- | --- |
| Decision Authority | Human retains final decision authority | AI drives decisions with selective human intervention |
| Role of AI | Assistive (analysis, recommendation, perception) | Active (decision-making and action execution) |
| Role of Human | Primary decision-maker | Supervisor, validator, or escalation authority |
| Control Structure | Human-centered | AI-centered with human oversight |
| Evaluation Focus | Human outcomes (usability, trust, decision quality) | Model/system performance (accuracy, reliability) |
| Risk Profile | Misinterpretation, over-reliance by human | Incorrect actions, bias, misalignment by AI |
| Intervention Pattern | Continuous human involvement | Selective, event-driven human involvement |
| System Objective | Improve human decision-making | Enable controlled automation with safeguards |

**Table 1. Comparison of AI-in-the-Loop and Human-in-the-Loop Systems**

## III. HITL AS AN INDEPENDENT SYSTEM COMPONENT

### A. Evolution of Application Architecture

Enterprise software systems have historically evolved through progressive separation of concerns. Early systems were implemented as monolithic applications in which data management, business logic, workflow control, organizational structure, and user interaction were tightly coupled within a single system boundary. While this approach simplified initial development, it limited scalability, reuse, and adaptability.

| Era | Separation |
| --- | --- |
| Pre-1970s | Monolithic systems (logic, data, workflow, roles, UI combined) |
| 1970s | Data separated (databases) |
| 1980s | Workflow/process separated |
| 1990s | Roles and organization separated |

Table 2. Enterprise Software Systems Have Evolved Progressive Separation of Concerns

Subsequent architectural advances introduced systematic separation across these dimensions. Data management was externalized through database systems, enabling persistence and independent optimization. Workflow and process management were later separated into dedicated engines, allowing execution logic to be defined, modified, and monitored independently of application code. Organizational structure and role management were further decoupled through models that allow dynamic resolution of roles and responsibilities at runtime. Prior work on Organization Model Management (OMM) demonstrated that separating organizational logic from application workflows improves flexibility, maintainability, and alignment with evolving enterprise structures [4, 5].

These developments share a common pattern: system components that were once embedded within application logic are progressively externalized into reusable, independently managed subsystems. This pattern enables modularity, standardization, and cross-application consistency.

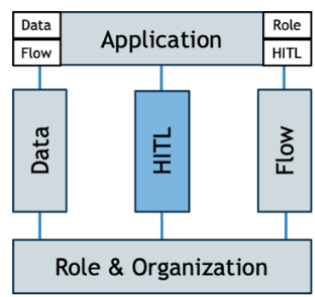

Figure 1. Decoupling HITL Component from the Application To Ensure Separation of Duty, Auditability, Safety, Security, Standardization, Scalability, Flexibility, Code Reuse, and Ease of Adoption

### B. Limitations of Embedded HITL Mechanisms

In current agent-based systems, Human-in-the-Loop functionality is typically implemented within application workflows. Human approval steps, escalation conditions, and role assignments are defined as part of the workflow logic, often using conditional rules or hard-coded decision points. While this approach is sufficient for isolated use cases, it introduces several limitations in multi-agent and enterprise-scale environments.

First, embedding HITL logic within applications reduces reusability. Each application must independently define how and when human intervention occurs, leading to duplicated logic and inconsistent behavior across systems. Second, governance policies such as approval thresholds, escalation rules, and audit requirements are difficult to enforce uniformly across the organization when they are distributed within multiple workflows. Third, this coupling limits the ability to observe and analyze human-agent interactions at a system level, since interaction data is fragmented across applications.

More importantly, embedding HITL within application logic constrains the evolution of agent autonomy. As agents transition from assistive to autonomous operation, the criteria for human involvement must be adjusted dynamically based on performance, risk, and context. Application-level implementations do not provide a systematic mechanism for managing this transition, making it difficult to scale autonomy in a controlled and measurable manner.

### C. *Decoupled HITL Architecture*

To address these limitations, we model HITL as an independent system component within the agent operating environment. In this architecture, human oversight is treated as a shared service that interacts with agent workflows through well-defined interfaces, rather than as embedded logic within individual applications.

The HITL component is responsible for managing the lifecycle of human interactions associated with agent decisions. This includes triggering human involvement based on predefined conditions, resolving appropriate human roles, capturing human input, and returning decisions or feedback to the agent workflow. By externalizing these responsibilities, the system establishes a consistent mechanism for integrating human oversight across different agents and applications.

This design aligns with the broader architectural pattern of separating cross-cutting concerns into dedicated subsystems. Similar to workflow engines or identity and access management systems, the HITL component provides a standardized interface that can be reused across domains. Agent workflows invoke the HITL component when human input is required, and continue execution based on the outcome of that interaction.

### D. *Interfaces and Execution Model*

In a decoupled architecture, interaction between agent workflows and the HITL component is mediated through explicit interfaces. At a minimum, these interfaces support three core operations: initiation of a human interaction request, delivery of contextual information required for decision-making, and receipt of human responses.

When an agent encounters a condition that requires human involvement, it generates a structured request that includes the relevant context, such as the current task state, proposed action, confidence level, and applicable constraints. The HITL component processes this request, determines the appropriate human participants based on role definitions, and routes the interaction through a suitable communication channel. Once a response is received, the result is returned to the agent, which resumes execution based on the outcome.

This execution model separates decision flow from interaction handling. Agent workflows remain focused on task execution, while the HITL component manages human engagement, communication, and response tracking. As a result, the system maintains a clear boundary between autonomous processing and human oversight.

### E. *Implications for Governance and Autonomy*

Treating HITL as an independent system component enables consistent enforcement of governance policies across agent workflows. Approval thresholds, escalation conditions, and audit requirements can be defined centrally and applied uniformly, reducing variability in system behavior. In addition, interaction data can be collected in a unified manner, enabling analysis of human-agent collaboration patterns and supporting continuous improvement.

This architecture also provides a foundation for progressive autonomy. Because human interactions are mediated through a structured component, the system can monitor outcomes, identify patterns in human decisions, and adjust intervention criteria over time. For example, repeated human approval of similar actions may indicate that certain decisions can be safely automated, while frequent overrides may signal the need for additional constraints or model refinement.

By decoupling HITL from application logic, the system supports both scalability and adaptability. It allows new agent workflows to inherit existing governance mechanisms without reimplementation, and enables organizations to evolve their oversight strategies as agent capabilities improve.

## IV. DESIGN MODEL FOR HUMAN-IN-THE-LOOP INTEGRATION

The decoupled HITL architecture requires a structured model to determine how human involvement is incorporated into agent workflows. In practice, human intervention cannot be continuous, as this would negate the efficiency gains of automation. At the same time, insufficient oversight increases the risk of incorrect or unsafe actions. Effective integration therefore depends on selectively engaging human participants based on well-defined criteria.

This section defines four dimensions that govern the integration of human oversight within agentic systems: intervention conditions (WHEN), role resolution (WHO), interaction semantics (WHAT), and communication channel (WHERE). Together, these dimensions provide a systematic framework for managing human involvement in a manner that balances autonomy and control.

### A. *WHEN: Intervention Conditions*

The first dimension concerns the conditions under which human involvement is triggered. In agentic workflows, intervention should be event-driven rather than continuous, and should occur only when specific criteria are met. These criteria are typically associated with risk, uncertainty, or policy constraints.

Examples include actions with significant financial or operational impact, situations where model confidence falls below a defined threshold, or cases that fall outside known distributions. In regulated environments, intervention may also be required to satisfy compliance requirements. By defining explicit triggering conditions with satisfying criteria like a rubric, the system ensures that human attention is directed to cases where it provides the highest value.

The design objective is to minimize unnecessary intervention while maintaining acceptable risk levels. This requires that triggering conditions be both precise and adaptable, allowing adjustment as system performance improves over time. By learning from the triggering condition, we can introduce agent autonomy by enabling AI to observe when human should be involved.

### B. *WHO: Role Resolution*

The second dimension addresses the identification of appropriate human participants. Human involvement is not interchangeable; it depends on organizational roles, responsibilities, and domain expertise. As a result, the system must resolve which individual or group is responsible for a given decision at runtime.

This capability builds on prior work in organizational modeling, where roles are defined independently of specific workflows and resolved dynamically based on context . For example, financial approvals may require a manager within a specific reporting hierarchy, while compliance-related decisions may require a designated regulatory officer. In other cases, domain experts may be required to provide specialized judgment.

Dynamic role resolution enables flexibility and scalability. It allows the same workflow to operate across different organizational structures without modification, and ensures that responsibility is assigned consistently according to defined policies.

### C. *WHAT: Interaction Semantics*

The third dimension defines the type of human involvement required for a given interaction. Human participation in agent workflows can take multiple forms, ranging from direct control to passive observation. These forms must be explicitly defined to ensure consistent behavior across the system.

At one end of the spectrum, humans may be required to approve or reject an agent's proposed action, effectively serving as a decision gate. In other cases, humans may provide additional context or corrections that influence the agent's subsequent behavior. In lower-risk scenarios, human involvement may be limited to monitoring or audit without direct intervention.

Defining interaction semantics allows the system to distinguish between different levels of control and responsibility. It also enables the collection of structured feedback, which can be used to refine models and adjust intervention policies over time.

### D. *WHERE: Communication Channel*

The fourth dimension concerns how human interaction is delivered and completed. In enterprise environments, human participants operate across multiple communication channels, including dedicated application interfaces, messaging platforms, and email systems. The choice of channel affects response time, user experience, and overall efficiency.

The HITL component should support flexible routing of interaction requests based on context, company communication policy, urgency, and user preference. For example, time-sensitive approvals may be routed through real-time messaging platforms such as SMS or Slack, while lower-priority interactions may be handled asynchronously through email or system dashboards or AMP UI portal.

Separating channel selection from workflow logic ensures that communication strategies can evolve independently of application behavior. It also allows organizations to optimize human engagement without modifying agent workflows.

*E.     Integrated View*

These four dimensions (WHEN, WHO, WHAT, and WHERE) operate jointly to define the behavior of the HITL system. Timing determines when intervention occurs, role resolution determines who is involved, interaction semantics define what form the involvement takes, and communication channels determine how the interaction is executed. By explicitly modeling these dimensions, the system provides a consistent and extensible framework for integrating human oversight across agent workflows.

This structured approach enables the system to manage human involvement as a controllable variable rather than an ad hoc process. As a result, it supports both the safe deployment of agents in high-risk environments and the gradual reduction of intervention as reliability improves.

The model can be aligned with emerging industry standards for agent interoperability, such as the Agent-to-Agent (A2A) communication protocol. Within such frameworks, agent capabilities, request structures, and interaction metadata provide a natural substrate for representing human-in-the-loop control. The four dimensions defined above can be mapped to standardized constructs: intervention conditions can be expressed as part of request context and policy constraints, role resolution can be derived from identity and capability metadata, interaction semantics can be encoded in action types and response schemas, and communication channels can be handled through protocol-level routing and integration mechanisms. This alignment allows HITL behavior to be implemented in a consistent and interoperable manner across heterogeneous agent systems without requiring application-specific customization. This shows that our proposed HITL model can be treated as a protocol-level concern rather than an application-level feature.

```
{
    "$schema": "http: //json-schema.org/schema#",
    "definitions": {
       ...
    }
    "A2ARequest": {
       ...
    }
    "AgentCapabilities": {
       "description": "Defines capabilities supported by an agent.",
       ...
    }
    "WHEN": "Condition or criteria to loop in the human."
    "WHO":  "Organization roles or user ID."
    "WHAT": "Approve/Reject/Modify/Defer | Enrich context | Notification"
    "WHERE":  "AMP | Slack | Email ..."
}
```

**Figure 2. The Independent HITL Model Enabling Interoperability With Agent Protocol**

## V.  AGENT INTERFACE TO THE HITL COMPONENT

The decoupled HITL architecture requires a well-defined interface between agent workflows and the HITL component. This interface enables agents to delegate decision control to the HITL system and to resume execution based on the outcome of human or automated evaluation. In practice, this interaction is implemented through a small set of RESTful interfaces that encapsulate the lifecycle of a HITL request.

The integration model follows a request–resolution pattern. An agent invokes the HITL component when a decision point satisfies predefined intervention conditions. The HITL component evaluates the request, determines whether human involvement is required, and manages the interaction until a resolution is reached. The agent then retrieves or receives the decision and continues execution accordingly.

*A. HITL Request Interface*

The first interface, denoted as `/api/hitl/request`, is used by the agent to initiate a HITL evaluation to determine if the requested agentic action should be approved. This call encapsulates the decision context and delegates control to the HITL component. The request includes structured information such as action fields, observed facts, and policy-related attributes required to evaluate intervention conditions. It also includes the criteria used to determine whether the action should be approved by a human or by the system. These criteria are collectively referred to as the *rubric*, which serves as the basis for decision outcomes such as *Approve, Reject, Defer*, or other gating decisions.

The caller may rely on the HITL component to retrieve relevant fields from enterprise systems through Model Context Protocol (MCP) calls and evaluate the criteria. Alternatively, the caller may provide these fields and criteria values directly for review by the approver.

Upon receiving the request, the HITL component evaluates whether the condition for human involvement are satisfied. If human intervention is required, the system resolves the appropriate participants, determines the interaction semantics, and routes the request through the selected communication channel. Otherwise, the system may return an automated decision without creating a human work item.

In addition to decision context, the request may optionally include a callback endpoint. When provided, this endpoint allows the HITL component to asynchronously deliver the HITL decision result to the caller. This enables event-driven integration and removes the need for repeated polling.

*B. Decision Retrieval Interface*

After initiating a HITL request, the agent must obtain the decision outcome before proceeding. The second interface, denoted as `/api/hitl/get-decision`, provides a mechanism for retrieving this result. This interface is typically implemented as a polling operation in which the agent periodically queries the HITL component using a unique request or instance identifier.

The response includes the current status of the request and, upon completion, the resolution of the decision along with metadata indicating whether the decision was produced by a human participant or an automated process. This approach allows the agent to remain stateless with respect to human interaction handling, delegating coordination responsibilities to the HITL component.

Polling provides a simple and robust integration model, particularly in environments where asynchronous callbacks are not supported or where network constraints limit inbound communication.

*C. Execution Modes*

The HITL integration supports two execution modes for receiving decision outcomes: polling and callback. In the polling mode, the agent repeatedly invokes the decision retrieval interface until the request reaches completion. In the callback mode, the agent supplies an external callback endpoint as part of the request, and the HITL component delivers the result asynchronously once available. The callback mode requires the caller to expose a reachable REST endpoint so that the HITL component can return the result. In contrast, the polling mode does not require inbound connectivity and can be used even when the agent operates in environments without a public URL or domain name, such as a local machine or a restricted network deployment.

These two modes represent alternative mechanisms for integrating HITL into agent workflows. Polling emphasizes simplicity and compatibility across environments, while callback-based delivery improves efficiency and responsiveness by eliminating repeated queries. Supporting both modes allows the HITL component to operate across a wide range of system architectures and deployment constraints.

## VI. SUMMARY AND CONTRIBUTION

This paper addresses the gap between conceptual frameworks for safe AI agents and their system-level implementation. While prior work established the importance of human oversight for transparency, accountability, and trustworthiness, existing approaches typically embed Human-in-the-Loop (HITL) mechanisms within application workflows. This design limits reuse, introduces inconsistencies in governance, and does not scale effectively in multi-agent environments.

To address this limitation, this paper presents the HITL model as an independent system component within the agent operating environment. By decoupling human oversight from application logic, the proposed architecture enables consistent integration of human intervention across heterogeneous multi-agent workflows. The component is defined through explicit interfaces and an execution model that separates agent decision flow from human interaction handling, allowing governance policies to be applied uniformly and managed centrally.

In addition, the paper introduces a structured design model for HITL integration based on four dimensions: intervention conditions, role resolution, interaction semantics, and communication channel. These dimensions provide a systematic basis for determining *when* human involvement is required, *who* should be involved, *what* form the interaction should be taken, and *where* it should be executed. This model allows human oversight to be treated as a controllable and observable aspect of system behavior rather than an application-specific construct.

The proposed approach also supports alignment with emerging agent communication standards. By mapping HITL control to protocol-level constructs, the design enables interoperability across agent systems and allows human oversight to be implemented as part of the agent interaction layer rather than as embedded workflow logic. This suggests that HITL can be treated as a protocol-level concern in agentic workflow architectures.

From a system perspective, the decoupled HITL component provides a foundation for progressive autonomy. Structured interaction data enables the analysis of human decisions and supports the adjustment of intervention policies over time. This allows agent systems to reduce human involvement in low-risk, well-understood scenarios while preserving oversight in high-risk or uncertain cases.

Overall, the contributions of this paper are threefold. First, it introduces an architectural model that externalizes HITL as an independent and reusable system component. Second, it defines a formal design model for integrating human oversight into agent workflows. Third, it establishes a pathway for aligning HITL with protocol-level standards to support interoperability and scalable governance in agentic systems.

The separation of HITL from application logic follows a broader pattern in software architecture. Data management, workflow execution, and organizational modeling were progressively externalized to improve scalability, flexibility, and code reuse. Applying the same principle, HITL can be externalized as a system-level component to support controlled and progressive autonomy in agentic systems. In this role, HITL functions not only as a safeguard, but as a control plane that governs the progression of increasing autonomy through structured human oversight and measurable outcomes.

### REFERENCES

[1] K. Lazaros, A.G. Vrahatis, S. Kotsiantis, *Human-in-the-Loop Artificial Intelligence: A Systematic Review of Concepts, Methods, and Applications,* Entropy, vol. 28, no. 4, p. 377, 2026. Available online: https://www.mdpi.com/1099-4300/28/4/377

[2] E. Cheng, J. Cheng, A. Siu, *Toward Safe and Responsible AI Agents: A Three-Pillar Model for Transparency, Accountability, and Trustworthiness, arXiv:2601.06223,* Jan. 2026. Available online: https://arxiv.org/abs/2601.06223

[3] S. Natarajan, et al. *Human-in-the-loop or AI-in-the-loop? Automate or Collaborate?*, in Proc. AAAI Conference on Artificial Intelligence (AAAI-25). Mar. 2025.

[4] E. Cheng, *An Object-Oriented Organizational Model to Support Dynamic Role-Based Access Control in Electronic Commerce*, Journal of Decision Support Systems, Mar. 2001.

[5] E. Cheng and G. Loizou, *A Publish/Subscribe Framework: Push Technology in E-Commerce*, Proceedings of the 17th British National Conference on Databases. Exeter, UK, Jul. 2000.